\documentclass[12pt]{article}
\usepackage[margin=1in]{geometry}
\usepackage{times}
\usepackage{microtype}
\usepackage{xcolor}
\usepackage{hyperref}
\usepackage{url}
\usepackage{booktabs}
\usepackage{amsmath, amssymb}
\usepackage{graphicx}
\usepackage{caption}
\usepackage{subcaption}
\usepackage{array}
\usepackage{multirow}
\usepackage{natbib}
\usepackage{float}
\usepackage{CJKutf8}
\usepackage{enumitem}
\setlist{nosep, topsep=2pt, partopsep=0pt}

\definecolor{darkblue}{rgb}{0, 0, 0.5}
\hypersetup{colorlinks=true, citecolor=darkblue, linkcolor=darkblue, urlcolor=darkblue}

\newcommand{\cn}[1]{\begin{CJK}{UTF8}{gbsn}#1\end{CJK}}
\newcommand{\visionupper}{\textit{Vision-50\%}}

\newcommand{\visionpartial}{\textit{Vision-80\%}}
\newcommand{\visionfull}{\textit{Vision-100\%}}

\newcommand{\textbased}{\textit{Index-based}}
\newcommand{\textmodel}{\textit{Index-based Model}}
\newcommand{\visualbased}{\textit{Visual-based}}
\newcommand{\visualmodel}{\textit{Visual-based Model}}

\title{Hot-Start Chinese Language Modeling:\\Visual Glyphs Accelerate Sample-Efficient Learning}

\author{
  Shuyang Xiang\thanks{vanillaxiangshuyang@gmail.com} \\
  Independent Researcher
  \and
  Hao Guan\thanks{guanhan032@gmail.com} \\
  Institute of Software, Chinese Academy of Sciences
}

\date{}

\begin{document}
\maketitle

\begin{abstract}
In this work,  we study whether rendering Chinese characters as visual glyph images---rather than discrete token IDs as mainstream LLMs do, providing an inductive bias for character-level language modeling.
Our central finding gives a double-edged insight: visual inputs produce a pronounced \textbf{hot-start effect}, more than doubling early-stage accuracy within the first epoch (at 0.4\% of total training steps) (12.3\% visual inputs vs.\ 5.8\% index-based baseline), yet both approaches converge to essentially identical final accuracy ($\sim$39\%).
This pattern holds across resolutions as low as $8{\times}8$ pixels, partial cropping up to 50\%, and model scales from 110M to 1.78B parameters.
The mechanism we identify is that glyph rendering \emph{pre-encodes} radical-based structure into embedding space before any training (cosine similarity 0.27 vs.\ 0.002 for random embeddings), enabling faster alignment but not higher final capacity.
Our results clarify both the promise and fundamental limitation of visual representations as inductive biases for Chinese language modeling.
\end{abstract}

% ========== 1. Introduction ==========
\section{Introduction}

A natural hypothesis in Chinese NLP is that the rich visual structure of Chinese characters, including stroke configurations, radical components, spatial layouts, should provide meaningful inductive bias for language models.
Unlike alphabetic languages such as English where subword tokenization captures morphological composition, standard character-level Chinese Language Models (LMs) assign arbitrary integer IDs to characters, generally discarding all sub-character structure.
For example, characters sharing the same radical \cn{扌} (hand), such as \cn{打} (hit), \cn{拍} (pat), and \cn{拉} (pull), exhibit similar left-right structure; characters like \cn{灭} (extinguish) and \cn{火} (fire) differ in top-bottom composition where the former symbolizes ``cover the fire (the latter)''. These structural regularities are entirely absent from randomly initialized embeddings.
Rendering characters as images would seem to restore this lost compositional information.

We test this hypothesis directly by replacing discrete token IDs with lightweight visual encoders that process rendered glyph images. We track learning dynamics carefully across training.
The result is instructive: visual inputs \emph{do} provide a useful prior---but only in the early-training regime.
Despite a striking hot-start advantage (2--3$\times$ higher accuracy within the first 1\% of training steps), visual and index-based models converge to identical final performance.

This paper contributes a careful characterization of \emph{when} and \emph{why} visual structure helps, and equally importantly, \emph{why it stops helping}.
We believe this finding is more informative than a simple positive result: it suggests that character visual priors reduce sample complexity for early learning, but the information bottleneck ultimately lies in distributional co-occurrence statistics that both representations must learn from data.

\begin{figure*}[t]
    \centering
    \includegraphics[width=0.9\textwidth]{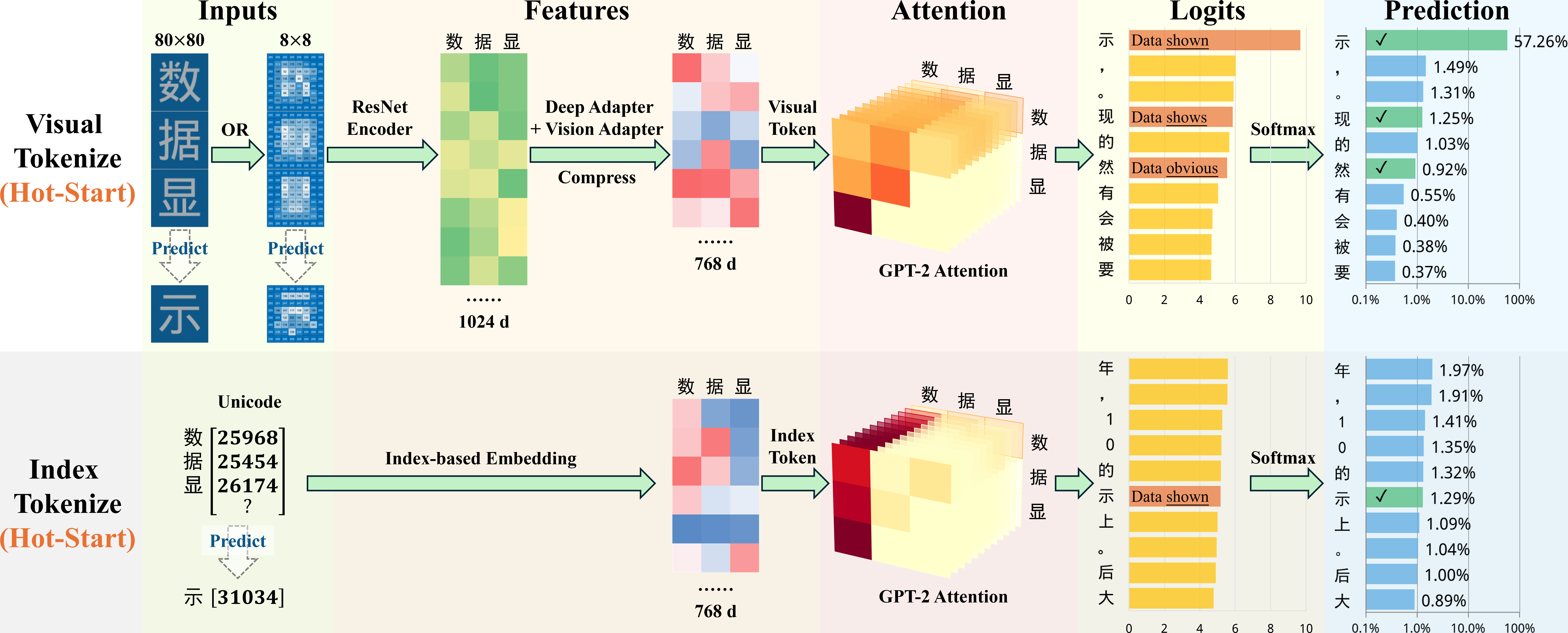}
    \captionsetup{font=footnotesize}
    \caption{
    Visual vs.\ \textbased{} pipelines for predicting the final character in ``\cn{数据显示}''.
    Visual inputs ($8\times8$) show a clear \emph{hot-start} advantage at 0.5\% training steps.
    }
    \label{fig:architecture-and-comparison}
\end{figure*}

\paragraph{Research Questions.} We investigate:

\noindent\textbf{RQ1 (Early-Stage Dynamics).} Do visual glyphs accelerate early learning compared to \textbased{} tokens input?

\noindent\textbf{RQ2 (Visual Sufficiency).} Can extremely low-resolution glyphs alone support competitive character prediction?

\noindent\textbf{RQ3 (Spatial Robustness).} Are \visualmodel{}s robust to degradation such as partial cropping?

\noindent\textbf{RQ4 (Resolution Sensitivity).} How does performance vary with image resolution?

\paragraph{Our Contributions.}
\begin{itemize}
\item We apply a vision-token formulation, replacing discrete character IDs with lightweight visual embeddings for Chinese language modeling.
\item We identify a \emph{hot-start} effect: visual inputs significantly accelerate early learning, doubling low-data performance within the first epoch.
\item \visualmodel{}s maintain competitive accuracy under extremely low resolution or severe cropping, demonstrating robustness.
\item Systematic analysis confirms minimal visual cues provide an effective inductive bias, revealing sample-efficient learning dynamics of Chinese language models.
\item We provide an interpretability analysis of how visual features drive the \emph{hot-start} effect and enable sample-efficient learning.
\item We provide open access to all training and experimental implementations: \url{https://github.com/ShuyangenFrance/chinese-vision}
% TODO: replace with your real github link before uploading to arxiv
\end{itemize}

% ========== 2. Related Work ==========
\section{Related Work}
\label{sec:related}

Different from alphabetic languages relying largely on subword tokenization~\cite{NEURIPS2020_1457c0d6, Bommasani2021OnTO, bender-koller-2020-climbing, rust-etal-2021-good}, mainstream Chinese language models still process characters as discrete token IDs, discarding the rich visual structure inherent in logographic scripts. While effective for sequence modeling, this \textbased{} representation ignores semantic and structural cues embedded in glyphs---cues that humans naturally exploit for reading.

Our work builds on and differs from three lines of prior research: glyph-augmented Chinese representations, pixel-based language models, and multimodal models processing text as images. Unlike prior work, we focus on pure visual inputs and their effect on early-stage, sample-efficient learning.

\subsection{Glyph-Augmented Chinese Representations}

Prior studies incorporate visual glyph features with token IDs~\cite{broscheit2018learning, Meng2019glyce, liu2020chinese, Sun2021ChineseBERT, zhu-etal-2023-glyph}. These methods still keep discrete token IDs as the primary representation while visual features serve as supplementary signals. We replace token IDs completely with visual inputs instead, isolating the inductive effect of glyphs on learning dynamics, especially in early-stage and low-resource regimes.

\subsection{Pixel-Based Language Models}

Pixel-based models treat text as images, performing masked reconstruction (PIXEL~\cite{rust2023pixel}, multilingual extensions~\cite{kesen2025multilingual}) or autoregressive pixel generation (PIXAR~\cite{ai2024pixar}). Some approaches apply pixel-based inputs to downstream tasks including machine translation, domain adaptation~\cite{salesky2021robust, salesky2023multilingual}, etc. These models operate in a visual-out path, predicting pixels rather than linguistic tokens. In contrast, our work applies a ``visual-in, token-out'' architecture, predicting next-character tokens from visual glyphs. This allows us to isolate how visual structure guides learning dynamics. 
\subsection{Multimodal Models with Visual Text}

Models such as CLIPPO~\cite{tschannen2023clippo}, DeepSeek-OCR~\cite{deepseek-ocr}, and Pix2Struct~\cite{Lee2022Pix2Struct} process text as images for OCR, document understanding, or retrieval. These systems prioritize downstream task performance or transcription accuracy. Rather than recognizing characters, we investigate whether visual forms alone suffice next token prediction. In addition, we focus on hot-start behavior and sample-efficient learning---an aspect largely unexplored in prior multimodal studies.

\subsection{Positioning This Work}

The work listed above demonstrate that visual features improve
 Chinese representations, that pixels can serve as language modeling targets, and that multimodal models can process visual text. It remains unexplored whether pure visual inputs produce systematically different \emph{learning dynamics} in an autoregressive setting---and if so, why. To our knowledge, no prior work systematically analyzes how visual structure influences learning dynamics in language modeling, nor isolates the hot-start phenomenon across different resolutions, cropping conditions, and model scales. We address this gap by studying autoregressive next-character prediction from pure visual inputs, with a particular focus on early-stage, sample-efficient learning.

% ========== 3. Methodology ==========
\section{Methodology}
\label{sec:method}

We study the role of visual glyphs in Chinese language modeling by comparing \textbf{\visualbased{} inputs} against standard \textbf{\textbased{} character IDs},  investigating the inductive effect of glyph structure.

\subsection{Model Architecture}

As illustrated in Figure~\ref{fig:architecture-and-comparison}, the only difference between our two paradigms lies in their input representations:

\begin{itemize}
    \item \textbf{\textbased{} path:} each character is assigned a unique index from a Chinese vocabulary  and mapped to a \emph{randomly initialized}, learnable embedding vector. No subword segmentation or pre-trained embeddings are used here; each character is treated as an atomic unit.
    \item \textbf{\visualbased{} path:} each character is rendered as a grayscale image and passed through a \textbf{lightweight ResNet encoder}~\cite{he2016deep} with an adapter module~\cite{Meng2019glyce}, then projected into the same embedding space.
\end{itemize}

All other parameters are identical so that any performance differences reflect \textbf{input representation only}. This guarantees that we can attribute \textbf{early-stage learning advantages (hot-start)} to visual structure rather than model capacity. See Appendix~\ref{app:efficiency} for detailed encoder ablations and visualizations.

\subsection{Visual Input Design}

To understand how minimal visual cues support prediction, we render characters at multiple resolutions: $80\times80$ down to $8\times8$ pixels. We test three \textbf{partial visibility} conditions:

\begin{itemize}
\item \visionfull{}: full glyph with natural margins
\item \visionpartial{}: only top 80\% of the character retained
\item \visionupper{}: only top 50\% of the character retained
\end{itemize}

This design draws inspiration from human reading: characters remain recognizable even when partially cropped. Observing robust prediction under cropping suggests the model \textbf{learns structural cues rather than relying on full OCR-style recognition}. Example glyph crops at different resolutions are shown in Appendix~\ref{app:cropping}.

\subsection{Training Objective and Setup}

Models are trained to predict the next character using \textbf{standard cross-entropy loss}. Gradients flow through the vision encoder, allowing visual features to adapt to the language modeling objective. To further highlight early-stage learning, we adopt a curriculum that gradually increases sequence length per epoch to clearly observe the hot-start effect. We provide full details of dataset, hyperparameters, curriculum, and optimization in Appendix~\ref{app:setup} for brevity.

Notably, even our simplest \visualmodel{} has fewer parameters than the index-based baseline (Appendix~\ref{app:efficiency}).

% ========== 4. Experiments ==========
\section{Experiments and Results}
\label{sec:results}

\subsection{Experimental Setup}

We evaluate three input configurations: the \textmodel{} baseline with standard token IDs, \visionfull{} with full character images, and \visionpartial{} with partial crops (top 80\% / 50\%). We set resolutions ranging from $4\times4$ to $80\times80$ pixels to control visual information density.

In our main experiments, all models are trained on THUCNews~\cite{thuctc2016} (100K sequences, 12.8M characters, length 128). We adopt a quadratic curriculum where sequence count grows as $5000 + 918.37e + 18.74e^2$ per epoch, with a 5K validation set. The decoder follows GPT-2-small~\cite{radford2019language} pre-trained on UER~\cite{zhao-etal-2019-uer}.

Full dataset details, hyperparameters, and additional decoder architectures are in Appendix~\ref{app:dataset} and~\ref{app:hyperparams}. We provide generalization experiments on Chinese Wikipedia and larger decoders in Appendix~\ref{app:experiments}.

\subsection{RQ1: Early-Stage Dynamics (Hot-Start)}
\label{sec:rq4}

\begin{figure}[t]
    \centering
    \includegraphics[width=0.9\columnwidth]{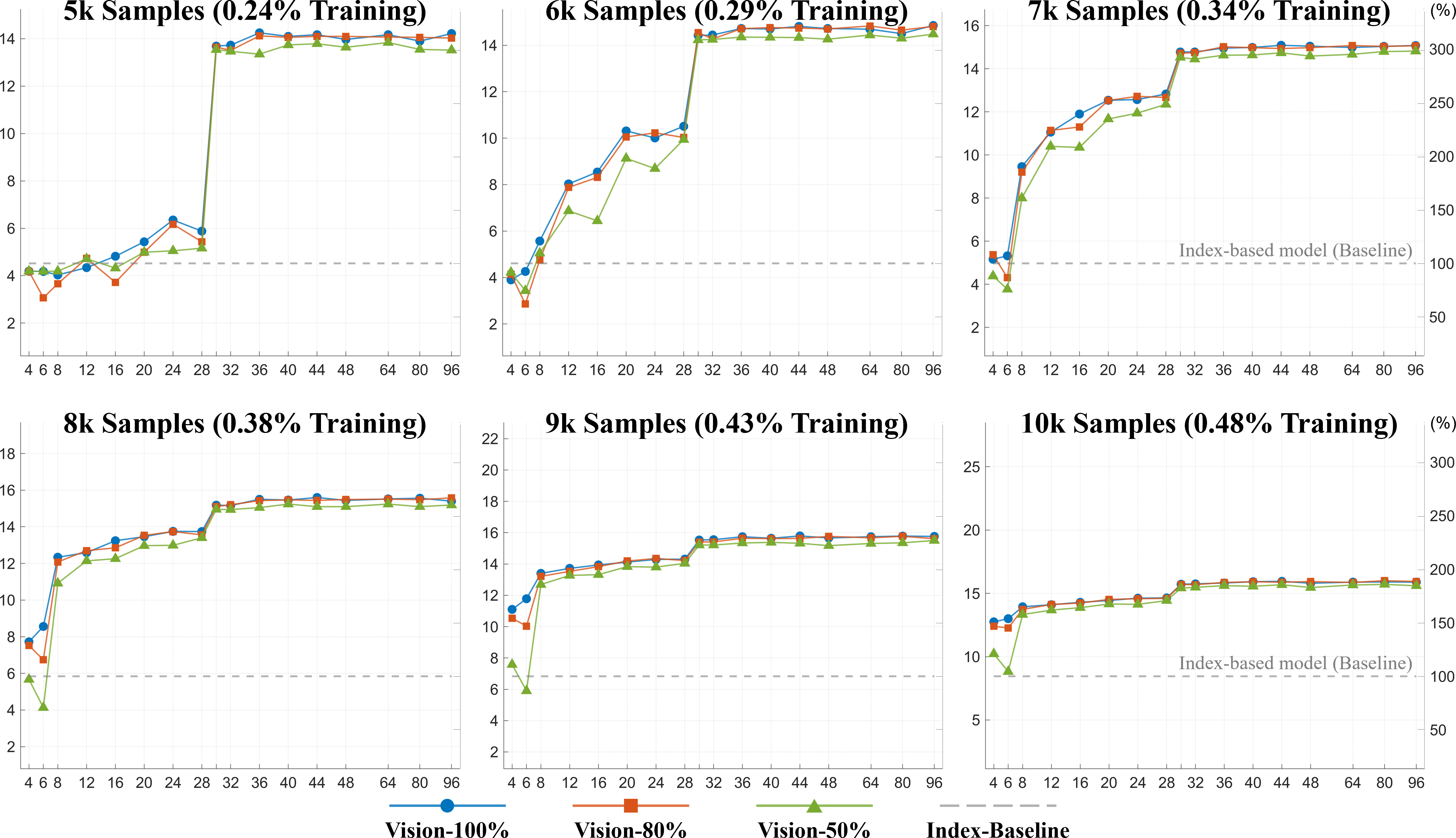}
    \captionsetup{font=small}
    \caption{Early-stage validation accuracy across image resolutions (5k--16k samples). The dashed line denotes the \textbased{} baseline.}
    \label{fig:hotstart-dynamics-selected}
\end{figure}

Figure~\ref{fig:hotstart-dynamics-selected} shows validation accuracy during early training (5k--16k samples), within the first 1\% of training steps. \visualmodel{}s consistently outperform the \textbased{} baseline across all early checkpoints.

At 4,096 samples (0.2\% of training steps), the \textbased{} baseline achieves only 4.30\% accuracy. In contrast, the $40\times40$ \visionfull{} already reaches 13.06\%, a $3\times$ improvement. At 8,200 samples (0.4\%), the $8\times8$ model reaches 12.34\%, doubling the baseline's 5.84\%. Higher-resolution inputs reach their hot-start phase earlier, suggesting richer visual detail accelerates structural extraction. Full trajectories are in Appendix~\ref{app:hotstart}.

This effect is robust across settings. On Chinese Wikipedia 2019 (zhwp2019), visual inputs achieve 8.88\% vs.\ text's 5.30\% at 8K samples. When scaling to a 1.78B-parameter model (DeepSeek-R1-Distill-Qwen-1.5B), \visualbased{} inputs again lead early under both repeated and incremental data regimes.

\textbf{These results demonstrate the hot-start effect}: visual structure provides an inductive bias that enables meaningful learning from extremely limited data, allowing models to converge faster within the first epoch.

\subsection{RQ2: Visual Sufficiency}
\label{sec:main-result}

The dataset contains over 5,500 distinct characters, so random guessing yields $\sim$0.02\% accuracy and a unigram baseline achieves $\sim$2\%. Against this accurancy, both visual and \textbased{} representations reach $\sim$39\% accuracy.

Table~\ref{tab:accuracy-ppl} shows that even at $8\times8$ resolution, the \visionfull{} achieves 39.21\%, matching the \textbased{} baseline (39.10\%). Therefore, \emph{coarse visual structure alone is sufficient} for next-character prediction.

\subsection{RQ3: Resolution and Spatial Robustness}
\label{sec:rq3}

At $4\times4$, accuracy drops to 29.70\%, indicating insufficient structure. However, performance saturates quickly: raising to $8\times8$ already recovers full accuracy, while higher resolutions provide only marginal gains. Fine-grained details contribute little beyond basic structural cues.

Robustness to spatial occlusion is striking: at $8\times8$, retaining only the top 80\% (\visionpartial{}) achieves 39.18\%, and even the top 50\% (\visionupper{}) reaches 38.63\%. We relate this to a ``toast-center'' effect where central strokes carry denser structural information (Appendix~\ref{app:toast-effect}).

Interestingly, higher resolutions beyond $8{\times}8$ provide no meaningful accuracy gains. However, robustness to 50\% occlusion suggests a possible application in document restoration: a model trained on Chinese character images could use glyph structure to predict or verify partially degraded characters in classical texts, even when large portions of the original strokes are lost.

Statistical analysis (DEFF=19.9, $\rho=0.15$) yields standard errors of 0.27--0.54\%, confirming differences across higher resolutions are not significant.

\begin{table}[h]
\centering
\tiny
\captionsetup{font=footnotesize}
\setlength{\tabcolsep}{1.5pt}
\renewcommand{\arraystretch}{0.8}
\begin{tabular}{lccccc}
\toprule
Mode & 4$\times$4 & 8$\times$8 & 20$\times$20 & 30$\times$30 & 80$\times$80 \\
 & Acc/PPL & Acc/PPL & Acc/PPL & Acc/PPL & Acc/PPL \\
\midrule
\visionfull{} & 29.70/85.33 & 39.21/46.59 & 39.16/45.83 & 39.14/48.73 & 39.03/49.41 \\
\visionpartial{} & 18.28/194.98 & 39.18/46.23 & 39.15/46.33 & 39.07/48.83 & 39.08/48.74 \\
\visionupper{}  & 2.10/2249.29 & 38.63/47.95 & 38.70/48.04 & 38.66/49.81 & 38.57/50.33 \\
\midrule
\textbased{} &  & 39.10/47.58 & & & \\
\bottomrule
\end{tabular}
\caption{Accuracy/PPL across resolutions. Performance saturates at $8\times8$, indicating sufficiency of coarse structure.}
\label{tab:accuracy-ppl}
\end{table}

\subsection{RQ4: Efficiency and Model Design}
\label{sec:efficiency}

Table~\ref{tab:consolidated-results} shows that the simplified \visionfull{} achieves the hot-start advantage with \textbf{33.5\% fewer parameters} (12.61M vs.\ 18.97M) and only 7\% additional FLOPs. Memory overhead is minimal (+1.3\%).

\begin{table*}[t]
\centering
\scriptsize
\captionsetup{font=small}
\setlength{\tabcolsep}{4pt}
\renewcommand{\arraystretch}{1.15}
\begin{tabular}{@{}llccccc@{}}
\toprule
\textbf{Category} & \textbf{Setting} & \textbf{Params} & \textbf{FLOPs} & \textbf{Throughput} & \textbf{Acc@8k} & \textbf{Final Acc} \\
\midrule
\multicolumn{7}{@{}l}{\textit{Baseline}} \\
Text & \textbased{} & 18.97M & 26.30G & 347.3$\pm$0.9 & 5.30\% & 39.10\% \\
\midrule
\multicolumn{7}{@{}l}{\textit{Vision Variants (8$\times$8)}} \\
\multirow{3}{*}{Vision}
& \visionfull{} (orig.) & 26.45M & 30.61G (+16\%) & 314.3$\pm$1.7 & 8.88\% & 39.21\% \\
& \visionfull{} (opt.)  & 22.32M & 29.56G (+12\%) & 306.9$\pm$0.7 & 8.75\% & 39.18\% \\
& \visionfull{} (simp.) & \textbf{12.61M} & 28.14G (+7\%) & 323.4$\pm$0.8 & 6.97\% & 39.19\% \\
\cmidrule(l){2-7}
& \visionpartial{} (80\%) & 12.61M & 28.14G (+7\%) & 323.4$\pm$0.8 & --- & 39.18\% \\
& \visionupper{} (50\%)   & 12.61M & 28.14G (+7\%) & 323.4$\pm$0.8 & --- & 38.63\% \\
\midrule
\multicolumn{7}{@{}l}{\textit{Ablations (vs \visionfull{} simp.)}} \\
Encoder      & ViT (8$\times$8)      & 14.23M & 29.87G (+13\%) & 298.5$\pm$1.2 & 6.12\% & 38.45\% \\
Training     & Frozen decoder        & 12.61M & 28.14G (+7\%)  & 323.4$\pm$0.8 & 4.89\% & 36.78\% \\
Architecture & No adapter            & 11.33M & 27.56G (+5\%)  & 335.2$\pm$0.6 & 5.21\% & 37.12\% \\
\midrule
\multicolumn{7}{@{}l}{\textit{Hot-Start (10K samples)}} \\
Early & \textbased{}             & 18.97M & 26.30G         & 347.3$\pm$0.9 & --- & 6.45\% \\
Early & \visionfull{} (simp.)    & 12.61M & 28.14G (+7\%)  & 323.4$\pm$0.8 & --- & 14.65\% \\
Early & \visionpartial{} (80\%)  & 12.61M & 28.14G (+7\%)  & 323.4$\pm$0.8 & --- & 14.70\% \\
Early & \visionupper{} (50\%)    & 12.61M & 28.14G (+7\%)  & 323.4$\pm$0.8 & --- & 14.38\% \\
\midrule
\multicolumn{7}{@{}l}{\textit{Statistical Validation}} \\
\multicolumn{7}{l}{Design effect DEFF=19.9 ($\rho=0.15$), $n_{\text{eff}}$=31.9K, 95\% CI width $\pm$0.537pp} \\
\bottomrule
\end{tabular}
\caption{Consolidated results at 8$\times$8 resolution. \visualmodel{}s match final accuracy while using fewer parameters, and show clear advantages in the hot-start regime.}
\label{tab:consolidated-results}
\end{table*}

At 8K samples, the simplified \visualmodel{} already outperforms the text baseline at 10K samples, confirming that the benefit of visual structure outweighs its modest computational cost. Full efficiency breakdown is in Appendix~\ref{app:efficiency}.

\subsection{Ablation Studies}

We conduct ablations at $8\times8$ to isolate contributing factors: (1) replacing ResNet with a ViT encoder yields only marginal changes, suggesting architectural choices are not critical; (2) freezing the decoder significantly degrades performance, confirming the importance of end-to-end training; (3) removing the adapter hurts accuracy, particularly during early training.

Notably, across all ablations, the hot-start effect persists, confirming that early acceleration arises from the visual structural prior rather than architectural specifics or parameter count.

\subsection{Transfer to Downstream Tasks}
\label{sec:ceval}

On C-Eval~\cite{NEURIPS2023_c6ec1844}, \visualmodel{}s outperform \textbased{} baselines by 17.5\% relative. Gains are strongest in STEM subjects (+9.8\% in physics, +8.5\% in mathematics, +14.6\% in chemistry). Notably, even at 10K samples, the \visionfull{} surpasses the \emph{fully trained} \textbased{} baseline overall (25.0\% vs.\ 22.3\%). In some subjects (HS chemistry, MS chemistry), the 10K-sample visual model outperforms its own fully trained version, suggesting extended training may introduce overfitting in the visual pathway for certain distributions. Full results are in Appendix~\ref{app:ceval}.

\subsection{Summary}

Three main findings emerge: (1) \textbf{Hot-start advantage}: visual inputs enable rapid early learning within the first epoch. (2) \textbf{Visual sufficiency}: coarse glyph structure alone supports competitive final performance. (3) \textbf{Robustness}: models remain stable under extreme downsampling and occlusion. Together, these results establish visual glyphs as a strong inductive bias for Chinese language modeling, particularly in low-resource regimes.

% ========== 5. Understanding the Hot-Start Effect ==========
\section{Understanding the Hot-Start Effect}
\label{sec:qualitative}

All analyses in this section focus on the $8\times8$ setting at the hot-start stage (with 10k training samples).

\subsection{Embedding Geometry as Structural Prior}
\label{sec:embedding-geometry}

We compare \textbased{} and \visualbased{} embeddings using L2-normalized Euclidean distance and cosine similarity across structural categories (indecomposable, left-right, top-bottom).

Table~\ref{tab:embedding-distances} shows that vision embeddings are consistently closer for visually similar characters: 1.2$\times$ smaller Euclidean distance and 30$\times$ higher cosine similarity. Within radicals (e.g., \cn{扌}, \cn{艹}), vision embeddings form coherent clusters (cosine $\sim$0.27), while index-based embeddings remain near zero.

This provides a geometric explanation for the hot-start effect: early in training, the visual model already operates in a representation space aligned with character structure.

Critically, this prior encodes \emph{visual} similarity rather than \emph{distributional} co-occurrence. Once the index-based model has seen enough data to learn co-occurrence patterns from scratch, the visual structural prior no longer confers a net advantage. This visual-distributional decoupling---characters sharing a radical appear in visually similar contexts but not necessarily similar linguistic contexts---may be a general property of logographic writing systems and explains why the advantage is limited to early training.

\textbf{(Full statistics and confidence intervals in Appendix~\ref{app:embedding-full})}

\begin{table}[t]
\centering
\scriptsize
\setlength{\tabcolsep}{2.5pt}
\renewcommand{\arraystretch}{0.9}
\begin{tabular}{@{}llll@{}}
\toprule
 & \textbf{\textmodel{}} & \textbf{\visualmodel{}} & \textbf{Ratio} \\
\midrule
\multicolumn{4}{l}{\textbf{Euclidean Distance}} \\
Similar pairs & 1.41 & 1.20 & 1.20:1 \\
\midrule
\multicolumn{4}{l}{\textbf{Cosine Similarity}} \\
Similar pairs & 0.01 & 0.28 & 30.4:1 \\
Radical \cn{艹} & 0.001 & 0.27 & --- \\
Radical \cn{扌} & 0.002 & 0.27 & --- \\
\bottomrule
\end{tabular}
\caption{Embedding structure at hot-start. Vision representations exhibit strong structural organization early in training.}
\label{tab:embedding-distances}
\end{table}

\begin{figure}[t]
    \centering
    \captionsetup{font=footnotesize}
    \includegraphics[width=0.9\columnwidth]{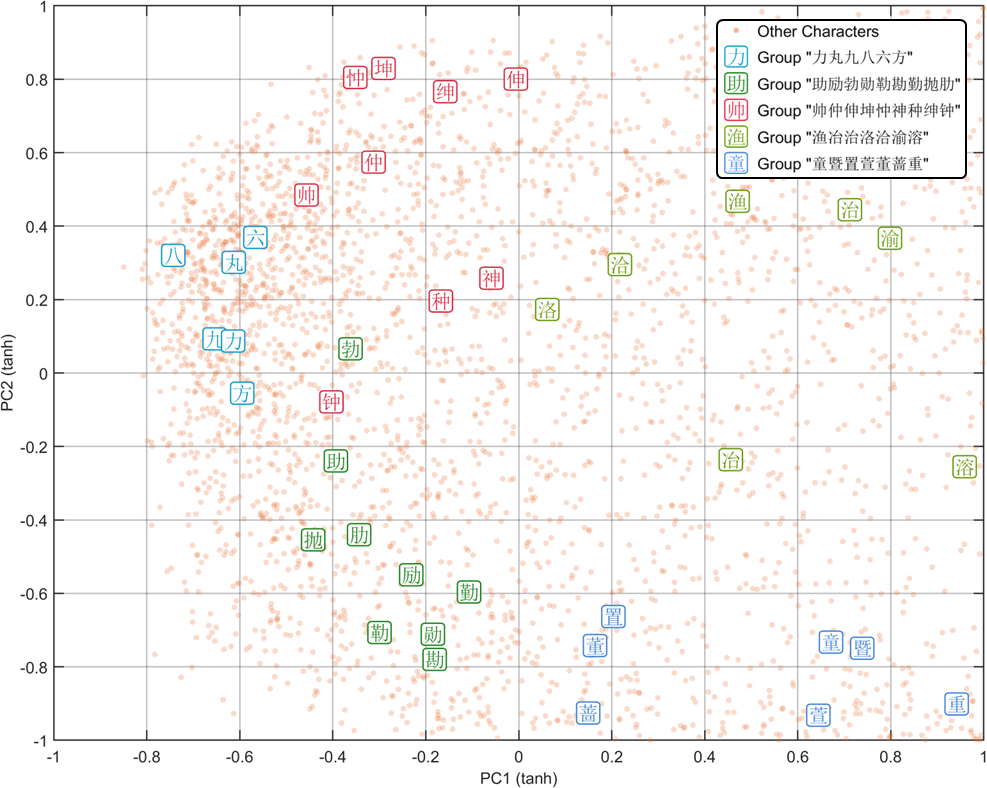}
    \caption{PCA of vision embeddings showing clustering of structurally similar characters before training.}
    \label{fig:pca-vision-only}
\end{figure}

\subsection{Early Discriminative Behavior}
\label{sec:behavior-similar}

We test confusable character pairs that differ in one or two strokes: \cn{土}/\cn{士}, \cn{人}/\cn{入}, \cn{日}/\cn{目}. At the hot-start stage, \visualmodel{}s consistently assign higher probability to the correct character, while index-based models often fail. For example, given ``\cn{下雨天鞋子上很容易沾上泥}'' (mud on shoes after rain), the visual model correctly prefers \cn{土} (earth/mud) over \cn{士} (scholar); the index model does not distinguish them. Across 8 such pairs, vision models show correct disambiguation in 6/8 cases vs.\ 3/8 for the index baseline.

\textbf{Interpretation:} structured embeddings directly translate into \emph{early discriminative ability}, explaining how vision models achieve higher accuracy with minimal data. Full results are in Appendix~\ref{app:similar-chars-full}.

\subsection{Distributed Pixel-Level Attribution}
\label{sec:gradient}

Via gradient-based attribution~\cite{zeiler_vis, vlm_grad}, we find that importance is distributed across the entire character rather than concentrated in specific regions. The model extracts predictive signals from multiple local regions rather than relying on specific strokes, making learning both robust and data-efficient. Full analysis is in Appendix~\ref{app:quantitative}.

\subsection{Summary: A Mechanism for Hot-Start}

Together, these analyses suggest a coherent mechanism:
\begin{itemize}
\item Visual inputs induce \textbf{structured embedding geometry} before training begins
\item This structure enables \textbf{immediate discriminative behavior}
\item Information is \textbf{distributed spatially}, ensuring robustness and efficiency
\end{itemize}

\subsection{Why Does Hot-Start Occur? Two Complementary Hypotheses}
\label{sec:hypotheses}

\paragraph{Hypothesis 1: Pre-Encoded Structural Prior.}
Glyph rendering \emph{pre-encodes} structural relationships into the input representation before any gradient-based learning. Characters sharing a radical (e.g., \cn{打}, \cn{拍}, \cn{拉}) begin with cosine similarity $\sim$0.27 at initialization vs.\ $\sim$0.002 for index embeddings. This warm-starts the representation space, reducing the representational distance the model must traverse during early optimization.

\paragraph{Hypothesis 2: Reduced Sample Complexity via Inductive Bias.}
Visual inputs reduce the number of training steps required to reach a given performance level. The structural prior constrains the hypothesis space, providing ``free'' information that index-based models must acquire from data alone. Evidence: the simplified visual encoder achieves the full hot-start advantage with 33.5\% fewer parameters; visual models surpass the fully trained index baseline on C-Eval at just 10K samples.

These hypotheses are reinforcing each other: the pre-encoded prior \emph{enables} the sample efficiency gain. They also explain why the advantage disappears at convergence---the prior encodes visual similarity, but prediction ultimately depends on distributional co-occurrence, which both representations must learn equally.

% ========== 6. Discussion ==========
\section{Discussion}
\label{sec:discussion}

\paragraph{Scope of our baseline.}
Our \textbased{} baseline uses \emph{randomly initialized, character-level} embeddings---not subword tokenization or pretrained representations. This controlled design isolates the effect of the visual prior, but means our conclusions apply specifically to this setting. The more practically relevant comparison against BPE-tokenized or pretrained baselines remains future work. We hypothesize the hot-start advantage would be substantially reduced against such baselines, as pretrained embeddings already encode rich distributional structure---which would clarify further that visual structure provides a scaffold for early learning, not a substitute for distributional information.

\paragraph{Implications for low-resource Chinese NLP.}
The hot-start effect is practically meaningful in data-scarce settings. At 10K samples, visual models already surpass the fully trained index baseline on C-Eval (+17.5\% relative). Visual glyph initialization could serve as a lightweight warm-start for low-resource fine-tuning scenarios, even when full training ultimately converges to the same ceiling.

\paragraph{Visual-distributional decoupling.}
An intriguing implication is that the visual structure of Chinese characters is \emph{not} strongly predictive of distributional co-occurrence in modern text. Characters sharing a radical are not necessarily similar in linguistic contexts---especially in news corpora. This decoupling may be a general property of logographic writing systems, worth investigating in its own right.

% ========== 7. Conclusion ==========
\section{Conclusion}
\label{sec:conclusion}

We identified a clear, reproducible hot-start effect from visual glyph inputs in Chinese language modeling, and equally clearly characterized its limits: the visual prior accelerates early convergence but does not change the final performance ceiling.
The underlying mechanism is geometric---pre-encoded radical-based embedding structure that warm-starts optimization---and requires only minimal visual information ($8{\times}8$ pixels, 33.5\% fewer parameters than the index baseline).

We summarize our findings through four research questions:
\begin{itemize}
    \item \textbf{RQ1 (Early-Stage Dynamics):} Visual models exhibit a strong \emph{hot-start}, reaching 12.3\% accuracy within the first 1\% of training steps---more than doubling the baseline (5.8\%).
    \item \textbf{RQ2 (Visual Sufficiency):} Even low-resolution inputs ($8\times8$) match \textbased{} performance (39.2\% vs.\ 39.1\%).
    \item \textbf{RQ3 (Resolution and Spatial Robustness):} Performance remains stable under extreme downsampling and cropping (top 50\% retained).
    \item \textbf{RQ4 (Efficiency):} Visual models achieve these gains with modest computational overhead and improved sample efficiency.
\end{itemize}

We hope this nuanced picture is more useful than a simple positive result: it clarifies exactly when visual representations of Chinese characters help, when they stop helping, and why. An intriguing direction for future work is whether the hot-start advantage relates to the \emph{compositionality} of visual representations---visual rendering may restore a form of sub-character ``atomicity'' through radicals, strokes, and spatial layout that character-level index embeddings for Chinese lack entirely.

\section*{Acknowledgements}
The authors used Claude (Anthropic) for assistance with English writing and \LaTeX{} formatting. All research content, experimental design, results, and conclusions are the authors' own.

\bibliography{refs}
\bibliographystyle{plainnat}

\appendix

% ========== Appendix A ==========
\section{Additional Quantitative Analysis}
\label{app:quantitative}

\subsection{Visual Importance Analysis}

To further investigate how visual feature importance is distributed, we analyze the central 70\% region of character images using the Center Mass Ratio (CMR) and attribution entropy.

\paragraph{Experimental Design.}
We evaluate models trained on sample sizes ranging from 1,024 to 16,441 sequences and resolutions from $4\times4$ to $96\times96$. We compute median and IQR of contrastive model error (CME) and entropy. Gap metrics quantify performance degradation when masking the top 5\%, 10\%, and 20\% of important regions.

\paragraph{Observations.}
\begin{itemize}
    \item \textbf{CME and Entropy Trends:} CME decreases for lower resolutions as sample size increases. Higher resolutions ($\geq16\times16$) show stable, positive CME. Entropy increases with resolution.
    \item \textbf{Gap Metrics:} Gap values increase with both sample size and resolution, suggesting diminishing returns at high sample sizes.
    \item \textbf{Resolution Dependence:} Low-resolution models show larger variance; higher-resolution models are more stable.
\end{itemize}

\begin{table}[ht]
\centering
\small
\begin{tabular}{ccccc}
\toprule
Image size & CME & CME IQR & Entropy & Gap (5\%/10\%/20\%) \\
\midrule
4  & $-$0.331 & 0.100 & 1.895 & 0.446 / 0.370 / 0.370 \\
6  & 0.272  & 0.022 & 2.227 & 0.493 / 0.446 / 0.446 \\
8  & 0.032  & 0.077 & 2.909 & 0.470 / 0.457 / 0.457 \\
12 & 0.187  & 0.042 & 3.468 & 0.533 / 0.452 / 0.452 \\
16 & 0.253  & 0.027 & 3.814 & 0.530 / 0.423 / 0.423 \\
20 & 0.221  & 0.031 & 4.298 & 0.570 / 0.506 / 0.506 \\
24 & 0.202  & 0.027 & 4.593 & 0.542 / 0.470 / 0.470 \\
28 & 0.173  & 0.038 & 4.853 & 0.541 / 0.465 / 0.465 \\
30 & 0.194  & 0.031 & 4.971 & 0.513 / 0.433 / 0.433 \\
32 & 0.232  & 0.021 & 5.043 & 0.539 / 0.448 / 0.448 \\
36 & 0.229  & 0.025 & 5.241 & 0.515 / 0.478 / 0.478 \\
40 & 0.225  & 0.029 & 5.495 & 0.495 / 0.443 / 0.443 \\
44 & 0.227  & 0.023 & 5.659 & 0.487 / 0.432 / 0.432 \\
48 & 0.209  & 0.030 & 5.786 & 0.485 / 0.427 / 0.427 \\
64 & 0.226  & 0.025 & 6.323 & 0.456 / 0.379 / 0.379 \\
80 & 0.226  & 0.026 & 6.775 & 0.384 / 0.331 / 0.331 \\
96 & 0.234  & 0.023 & 7.091 & 0.385 / 0.325 / 0.325 \\
\bottomrule
\end{tabular}
\caption{Internal metrics at sample size 10,248.}
\label{tab:app-visual-importance}
\end{table}

\subsection{Additional Accuracy Results}

\begin{table}[ht]
\centering
\small
\begin{tabular}{ccc}
\toprule
Image size & Training type & Top-5 (\%) \\
\midrule
8  & \visionfull     & 65.28 \\
8  & \visionpartial  & 65.26 \\
8  & \visionupper    & 64.78 \\
\midrule
20 & \visionfull     & 65.24 \\
20 & \visionpartial  & 65.24 \\
20 & \visionupper    & 64.81 \\
\midrule
30 & \visionfull     & 65.36 \\
30 & \visionpartial  & 65.35 \\
30 & \visionupper    & 64.87 \\
\midrule
80 & \visionfull     & 65.37 \\
80 & \visionpartial  & 65.36 \\
80 & \visionupper    & 64.90 \\
\bottomrule
\end{tabular}
\caption{Best full and partial training accuracy across image resolutions.}
\label{tab:app-full-partial}
\end{table}

\begin{table}[ht]
\centering
\small
\begin{tabular}{cccc}
\toprule
Image Size & \visionfull & \visionpartial & \visionupper \\
\midrule
4   & 27.42 & 26.76 & 23.46 \\
8   & 29.67 & 29.22 & 28.48 \\
16  & 30.35 & 30.06 & 29.25 \\
32  & 33.92 & 33.75 & 33.39 \\
48  & 34.35 & 34.27 & 33.56 \\
96  & 34.28 & 34.32 & 33.84 \\
\bottomrule
\end{tabular}
\caption{Hot-start Top-5 accuracy (\%) at 10,248 samples.}
\label{tab:app-hotstart-top5}
\end{table}

% ========== Appendix B ==========
\section{Training and Dataset Details}
\label{app:setup}

\subsection{Training Objective and Curriculum}

Models minimize standard cross-entropy loss:
\begin{equation}
\mathcal{L}_{\text{CE}} = -\frac{1}{N} \sum_{i=1}^{N} \sum_{t=1}^{T} \log P(c_{t+1}^{(i)} | I_1^{(i)}, \ldots, I_t^{(i)}),
\end{equation}
where $I_t^{(i)}$ are either visual features or token embeddings. We employ a quadratic curriculum where training sequences grow as $5000 + 918.37e + 18.74e^2$ per epoch.

\subsection{Dataset Details}
\label{app:dataset}
THUCNews~\cite{thuctc2016} is based on RSS feeds from Sina News (2005--2011), containing approximately 740,000 news articles. The corpus is UTF-8 encoded, with 100K sequences (12.8M character instances) split into fixed-length sequences of 128 characters. The validation set contains 5K fixed sequences.

\subsection{Full Hyperparameters}
\label{app:hyperparams}
\begin{itemize}
    \item \textbf{Model:} GPT-2-small (12 layers, 768 hidden, 12 heads), pre-trained on UER~\cite{zhao-etal-2019-uer}
    \item \textbf{Optimizer:} AdamW ($\beta_1=0.9$, $\beta_2=0.999$, $\epsilon=1\times10^{-8}$)
    \item \textbf{Learning rate:} $2\times10^{-4}$ with OneCycle scheduler (max $1.5\times10^{-3}$)
    \item \textbf{Batch size:} 128; \textbf{Weight decay:} 0.01; \textbf{Gradient clipping:} 1.0
    \item \textbf{Mixed precision:} FP16; \textbf{Early stopping:} patience 7 epochs
    \item \textbf{Resolution spectrum:} $4\times4$ through $96\times96$ (17 settings)
\end{itemize}

% ========== Appendix C ==========
\section{Full Experimental Results}
\label{app:experiments}

\subsection{Full Hot-Start Progression}
\label{app:hotstart}

\begin{table}[ht]
\centering
\scriptsize
\begin{tabular}{cccc}
\toprule
\textbf{Samples} & \textbf{Baseline} & \textbf{8$\times$8 Vision} & \textbf{40$\times$40 Vision} \\
\midrule
4,096  & 4.30\% & 4.19\% ($-$0.11\%) & 13.06\% (+8.76\%) \\
6,152  & 4.61\% & 5.57\% (+0.96\%)  & 14.70\% (+10.09\%) \\
8,200  & 5.84\% & 12.34\% (+6.50\%) & 15.46\% (+9.62\%) \\
10,248 & 8.45\% & 13.94\% (+5.49\%) & 15.92\% (+7.47\%) \\
12,296 & 9.87\% & 14.78\% (+4.91\%) & 16.21\% (+6.34\%) \\
14,344 & 11.23\% & 15.21\% (+3.98\%) & 16.45\% (+5.22\%) \\
16,441 & 13.33\% & 15.65\% (+2.32\%) & 16.68\% (+3.35\%) \\
\bottomrule
\end{tabular}
\caption{Full hot-start progression with difference percentages.}
\label{tab:app-hotstart-full}
\end{table}

\subsection{Full Wikipedia Validation Results}
\label{app:wikipedia}
On Chinese Wikipedia 2019 (zhwp2019):
\begin{itemize}
    \item At 8k samples: 8$\times$8 vision achieves 8.88\% vs.\ text's 5.30\%
    \item At 10k samples: 8$\times$8 vision achieves 14.65\% vs.\ text's 6.45\%
    \item Final accuracy: vision 32.4\% vs.\ text 32.1\%
\end{itemize}

\subsection{Full Efficiency Analysis}
\label{app:efficiency}
\textbf{Parameter breakdown:}
\begin{itemize}
    \item Text baseline: embedding (9.48M) + output (9.49M) = 18.97M
    \item Vision (orig.): ResNet (15.23M) + adapters (1.84M) + output (9.49M) = 26.45M
    \item Vision (opt.): encoder (11.10M) + adapters (1.84M) + output (9.49M) = 22.32M
    \item Vision (simp.): encoder (1.28M) + adapter (1.84M) + output (9.49M) = 12.61M
\end{itemize}

\textbf{Memory:} GPT-2 activations 7.58GB (94\%); vision encoder 0.27GB (3.3\%); total 8.06GB $\to$ 8.33GB (+1.3\%).

\subsection{Full Scalability Results}
\label{app:scalability}

\begin{table}[ht]
\centering
\begin{tabular}{lccc}
\toprule
Model & Regime & Acc@4k & Acc@5k \\
\midrule
Text & Repeated & 6.65\% & 9.57\% \\
Vision (8$\times$8) & Repeated & 10.58\% & 13.11\% \\
Text & Incremental & 6.63\% & 9.45\% \\
Vision (8$\times$8) & Incremental & 10.23\% & 13.01\% \\
\bottomrule
\end{tabular}
\caption{Hot-start on DeepSeek-R1-Distill-Qwen-1.5B (1.78B parameters).}
\label{tab:app-scalability}
\end{table}

Under repeated-data training, vision improves until epoch 8 (25.25\%) while text peaks at epoch 3 (19.48\%)---a 5.77pp advantage.

% ========== Appendix D ==========
\section{Full Embedding Statistics}
\label{app:embedding-full}

\begin{table}[H]
\centering
\scriptsize
\begin{tabular}{@{}llll@{}}
\toprule
\textbf{Category} & \textbf{Metric} & \textbf{\textmodel{}} & \textbf{\visualmodel{}} \\
\midrule
\multicolumn{4}{l}{\textit{Indecomposable (e.g., \cn{戊/戌})}} \\
 & Euclidean (95\% CI) & 1.42 [1.40,1.44] & 1.21 [1.09,1.34] \\
 & Cosine (95\% CI) & 0.008 [$-$0.04,0.05] & 0.26 [0.12,0.41] \\
\midrule
\multicolumn{4}{l}{\textit{Left-Right (e.g., \cn{扌})}} \\
 & Euclidean (95\% CI) & 1.41 [1.39,1.44] & 1.20 [1.09,1.31] \\
 & Cosine (95\% CI) & 0.002 [$-$0.03,0.04] & 0.27 [0.15,0.40] \\
\midrule
\multicolumn{4}{l}{\textit{Top-Bottom (e.g., \cn{艹})}} \\
 & Euclidean (95\% CI) & 1.41 [1.39,1.44] & 1.20 [1.09,1.32] \\
 & Cosine (95\% CI) & 0.001 [$-$0.03,0.04] & 0.27 [0.14,0.40] \\
\bottomrule
\end{tabular}
\caption{Full embedding statistics with 95\% confidence intervals.}
\label{tab:app-embedding-full}
\end{table}

\subsection{Full Results for Visually Similar Characters}
\label{app:similar-chars-full}

\begin{table}[H]
\centering
\scriptsize
\setlength{\tabcolsep}{3pt}
\begin{tabular}{@{}ccllcc@{}}
\toprule
\textbf{ID} & \textbf{Sentence} & \textbf{Candidates} & \textbf{Model} & \textbf{P(\%)} & \textbf{Choice} \\
\midrule
1 & \cn{下雨天鞋子上很容易沾上泥} & \cn{土/士} & Vision & 0.05/0.00 & \cn{土}\checkmark \\
  & & & Text & 0.00/0.01 & \cn{士}$\times$ \\
2 & \cn{他是一个边境战} & \cn{士/土} & Vision & 0.01/0.00 & \cn{士}\checkmark \\
  & & & Text & 0.00/0.01 & \cn{土}$\times$ \\
3 & \cn{这地板的材料是实} & \cn{木/本} & Vision & 0.00/0.23 & \cn{本}$\times$ \\
  & & & Text & 0.00/0.31 & \cn{本}$\times$ \\
4 & \cn{别忘了拿作业} & \cn{本/木} & Vision & 0.04/0.00 & \cn{本}\checkmark \\
  & & & Text & 0.04/0.00 & \cn{本}\checkmark \\
5 & \cn{昨天周六，今天是星期} & \cn{日/目} & Vision & 0.19/0.12 & \cn{日}\checkmark \\
  & & & Text & 0.00/0.01 & \cn{目}$\times$ \\
6 & \cn{这个广告牌很醒} & \cn{目/日} & Vision & 0.09/0.08 & \cn{目}\checkmark \\
  & & & Text & 0.03/0.00 & \cn{目}\checkmark \\
7 & \cn{这个房间非请莫} & \cn{入/人} & Vision & 0.04/0.02 & \cn{入}\checkmark \\
  & & & Text & 0.06/0.10 & \cn{人}$\times$ \\
8 & \cn{介绍一下，这位是我的爱} & \cn{人/入} & Vision & 8.63/0.00 & \cn{人}\checkmark \\
  & & & Text & 0.06/0.25 & \cn{入}$\times$ \\
\bottomrule
\end{tabular}
\caption{Full predictions for visually similar character pairs.}
\label{tab:app-similar-chars}
\end{table}

% ========== Appendix E ==========
\section{Gradient Attribution Maps}
\label{app:gradient}

\begin{table}[ht]
\centering
\small
\begin{tabular}{@{}lcc@{}}
\toprule
\textbf{Region} & \textbf{Avg. Intensity} & \textbf{Std. Dev.} \\
\midrule
Upper Half & 0.087 & 0.014 \\
Lower Half & 0.081 & 0.014 \\
Left Half & 0.085 & 0.016 \\
Right Half & 0.083 & 0.012 \\
Upper-Left Quadrant & 0.088 & 0.015 \\
Upper-Right Quadrant & 0.086 & 0.014 \\
Lower-Left Quadrant & 0.082 & 0.013 \\
Lower-Right Quadrant & 0.080 & 0.014 \\
Center 50\% Region & 0.092 & 0.012 \\
Peripheral 50\% Region & 0.078 & 0.016 \\
\bottomrule
\end{tabular}
\caption{Attention intensity statistics across character regions.}
\label{tab:app-region-stats}
\end{table}

% ========== Appendix F ==========
\section{Cropping Visualization}
\label{app:cropping}

\begin{figure}[H]
    \centering
    \captionsetup{font=small}
    \begin{subfigure}[b]{0.48\linewidth}
        \centering
        \includegraphics[width=0.7\linewidth]{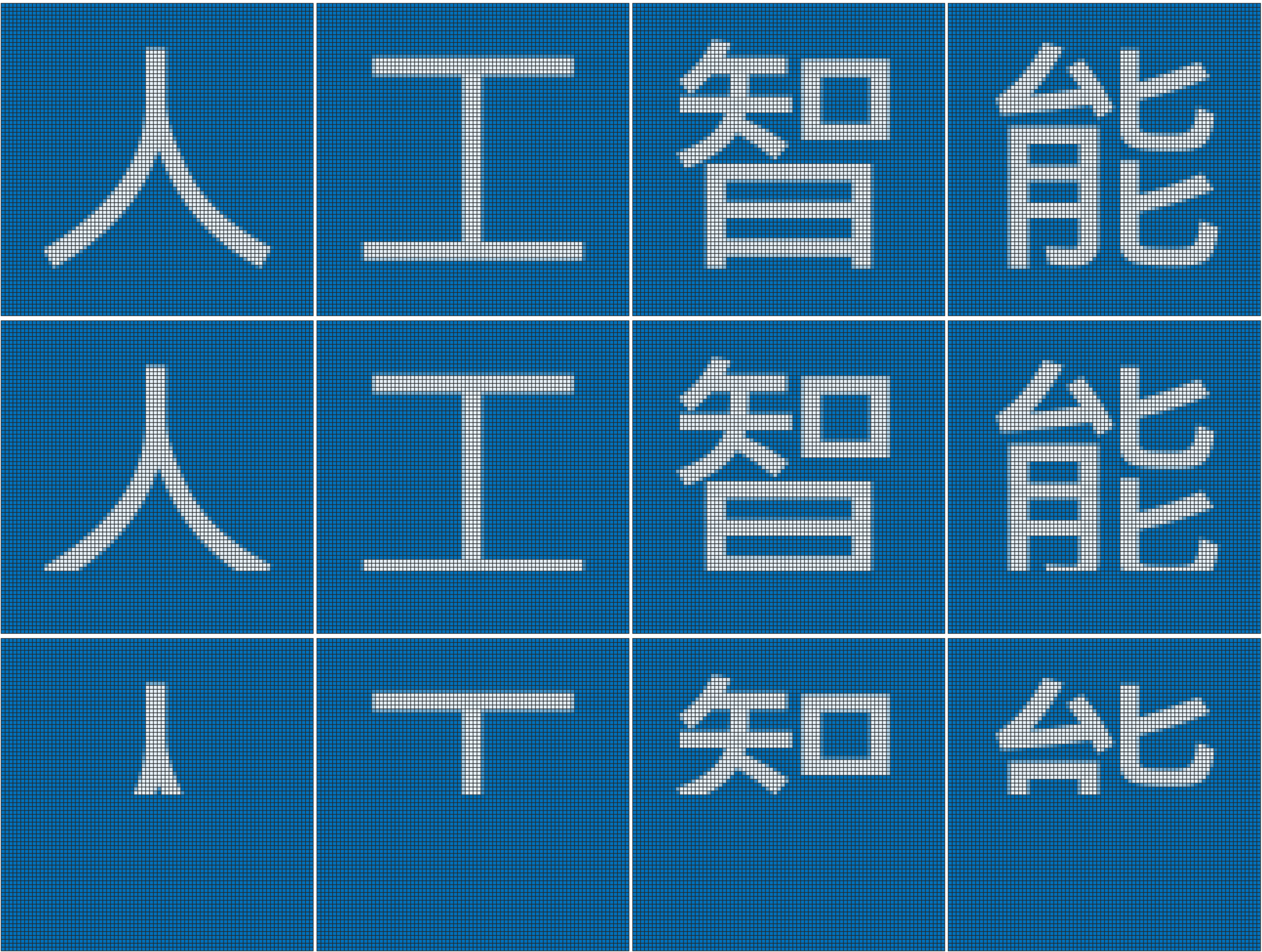}
        \caption{$80\times80$ pixels}
    \end{subfigure}
    \hfill
    \begin{subfigure}[b]{0.48\linewidth}
        \centering
        \includegraphics[width=0.85\linewidth]{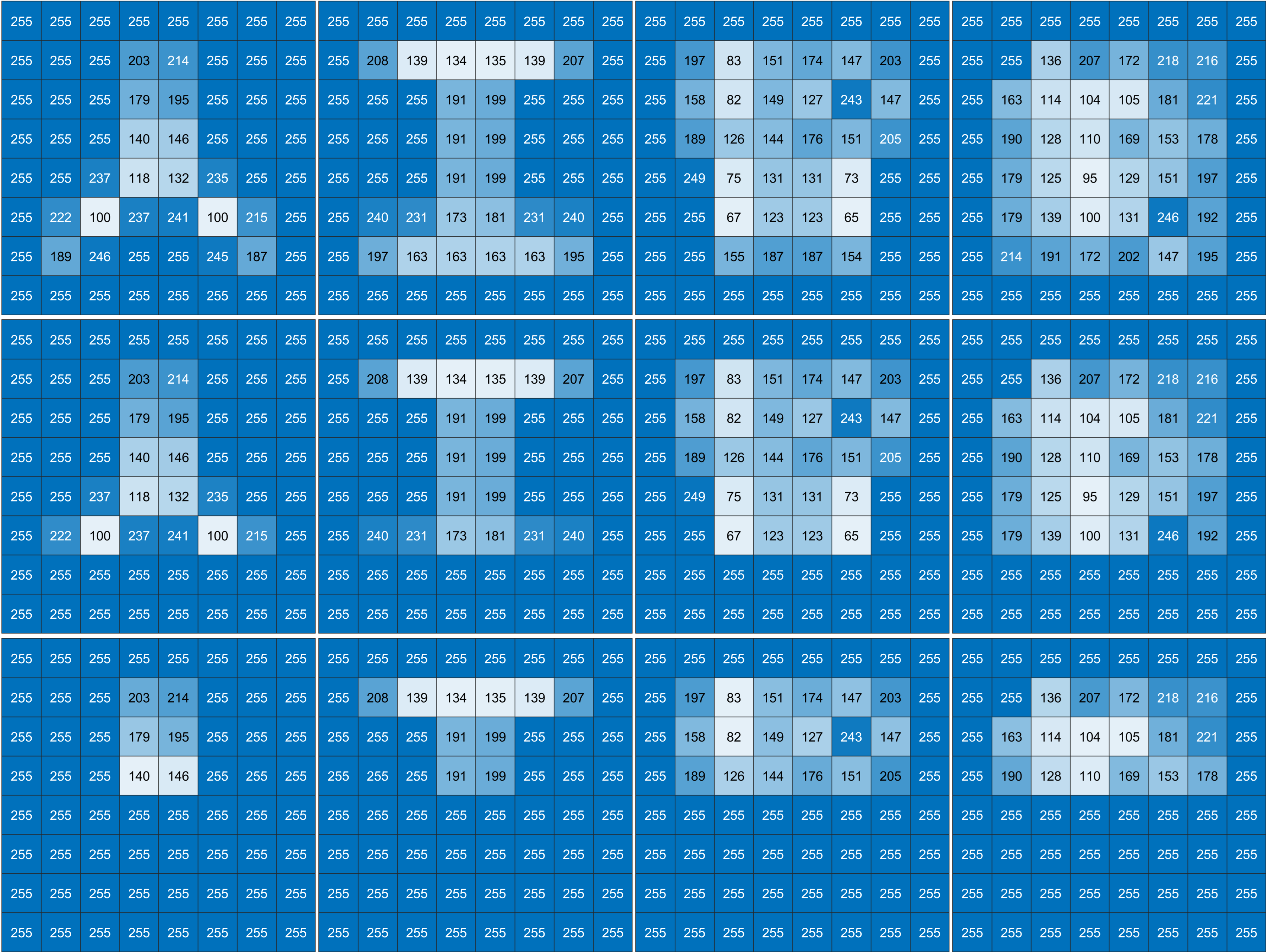}
        \caption{$8\times8$ pixels}
    \end{subfigure}
    \caption{``\cn{人工智能}'' at different resolutions and cropping levels. Even at $8{\times}8$ with 50\% cropping, core structure remains recognizable.}
    \label{fig:app-cropping}
\end{figure}

% ========== Appendix G ==========
\section{Toast-Center Effect Visualization}
\label{app:toast-effect}

\begin{figure}[H]
    \centering
    \includegraphics[width=0.8\columnwidth]{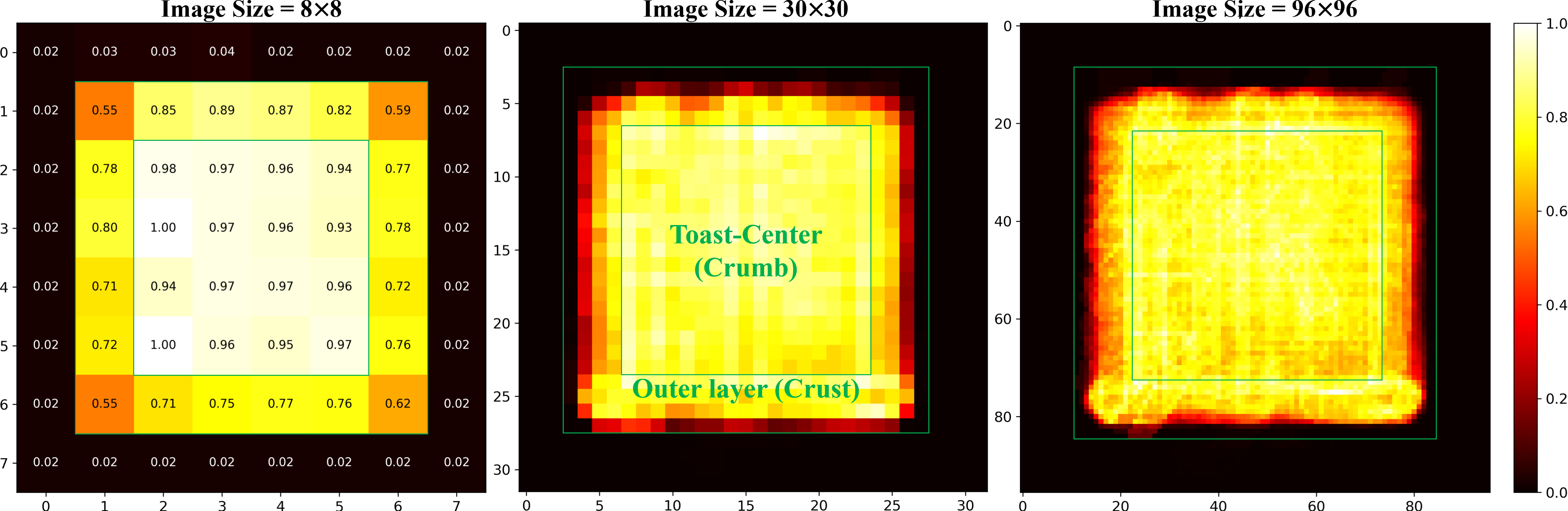}
    \captionsetup{font=small}
    \caption{``Toast-center effect'': central strokes (blue box) receive more attention than outer pixels (red box).}
    \label{fig:app-hotstart-toast}
\end{figure}

% ========== Appendix H ==========
\section{Full C-Eval Results}
\label{app:ceval}

\begin{table}[H]
\centering
\scriptsize
\begin{tabular}{lccc}
\toprule
\textbf{Subject} & \textbf{Text (full)} & \textbf{Vision (10k)} & \textbf{Vision (full)} \\
\midrule
advanced\_mathematics      & 25.4\% & 19.7\%          & \textbf{28.3\%} \\
discrete\_mathematics      & 24.2\% & 22.2\%          & \textbf{26.8\%} \\
high\_school\_chemistry    & 17.4\% & \textbf{32.0\%} & 25.6\% \\
high\_school\_chinese      & 25.8\% & 20.8\%          & \textbf{30.9\%} \\
high\_school\_mathematics  & 20.5\% & 21.7\%          & \textbf{28.3\%} \\
high\_school\_physics      & 17.1\% & 26.9\%          & \textbf{26.9\%} \\
logic                      & \textbf{27.9\%} & 27.0\% & 20.6\% \\
middle\_school\_chemistry  & 17.3\% & \textbf{31.9\%} & 21.1\% \\
middle\_school\_mathematics& 18.6\% & 22.0\%          & \textbf{27.1\%} \\
middle\_school\_physics    & \textbf{27.0\%} & 23.6\% & 26.4\% \\
probability\_and\_statistics& 23.5\% & 26.5\%         & \textbf{27.1\%} \\
\midrule
\textbf{Overall}           & 22.3\% & \textbf{25.0\%} & 26.2\% \\
\bottomrule
\end{tabular}
\caption{Full per-subject C-Eval accuracy. Vision (10k) surpasses fully trained text overall; vision (full) wins on 9/11 subjects.}
\label{tab:app-ceval-full}
\end{table}

\end{document}